%% file: main.tex
\documentclass[letterpaper]{article} 
\usepackage{aaai24}  
\usepackage{times}  
\usepackage{helvet}  
\usepackage{courier}  
\usepackage[hyphens]{url}  
\usepackage{graphicx} 
\usepackage{natbib}  
\usepackage{caption} 
\usepackage{algorithm}
\usepackage{algorithmic}

\usepackage{newfloat}
\usepackage{listings}
\DeclareCaptionStyle{ruled}{labelfont=normalfont,labelsep=colon,strut=off} 
\floatstyle{ruled}
\newfloat{listing}{tb}{lst}{}
\floatname{listing}{Listing}
\title{Multi-Energy Guided Image Translation with Stochastic Differential Equations for Near-Infrared Facial Expression Recognition}
\author{
    Bingjun Luo\textsuperscript{\rm 1},
    Zewen Wang\textsuperscript{\rm 1},
    Jinpeng Wang\textsuperscript{\rm 1},
    Junjie Zhu\textsuperscript{\rm 1},
    Xibin Zhao\textsuperscript{\rm 1},
    Yue Gao\textsuperscript{\rm 1}
}
\affiliations{
    \textsuperscript{\rm 1}BNRist, KLISS, School of Software, Tsinghua University\\
    luobingjun@gmail.com, \{wang-zw19,wjp21,zhujj18\}@mails.tsinghua.edu.cn, \{zxb, gaoyue\}@tsinghua.edu.cn
}

\usepackage{bibentry}

\usepackage{bm}
\usepackage{amssymb}
\usepackage{amsmath}
\usepackage{multirow}
\usepackage{booktabs}

\begin{document}

\input{math_commands.tex}

\maketitle

\begin{abstract}
Illumination variation has been a long-term challenge in real-world facial expression recognition(FER). Under uncontrolled or non-visible light conditions, Near-infrared (NIR) can provide a simple and alternative solution to obtain high-quality images and supplement the geometric and texture details that are missing in the visible domain. Due to the lack of existing large-scale NIR facial expression datasets, directly extending VIS FER methods to the NIR spectrum may be ineffective. Additionally, previous heterogeneous image synthesis methods are restricted by low controllability without prior task knowledge. To tackle these issues, we present the first approach, called for NIR-FER Stochastic Differential Equations (NFER-SDE), that transforms face expression appearance between heterogeneous modalities to the overfitting problem on small-scale NIR data. NFER-SDE is able to take the whole VIS source image as input and, together with domain-specific knowledge, guide the preservation of modality-invariant information in the high-frequency content of the image. Extensive experiments and ablation studies show that NFER-SDE significantly improves the performance of NIR FER and achieves state-of-the-art results on the only two available NIR FER datasets, Oulu-CASIA and Large-HFE.
\end{abstract}

\input{data/introduction}
\input{data/related_work}
\input{data/method}

\input{data/experiment}
\input{data/conclusion}

\bibliography{aaai24}

\end{document}

%% file: math_commands.tex

\newcommand{\figleft}{{\em (Left)}}
\newcommand{\figcenter}{{\em (Center)}}
\newcommand{\figright}{{\em (Right)}}
\newcommand{\figtop}{{\em (Top)}}
\newcommand{\figbottom}{{\em (Bottom)}}
\newcommand{\captiona}{{\em (a)}}
\newcommand{\captionb}{{\em (b)}}
\newcommand{\captionc}{{\em (c)}}
\newcommand{\captiond}{{\em (d)}}

\newcommand{\newterm}[1]{{\bf #1}}

\def\figref#1{figure~\ref{#1}}
\def\Figref#1{Figure~\ref{#1}}
\def\twofigref#1#2{figures \ref{#1} and \ref{#2}}
\def\quadfigref#1#2#3#4{figures \ref{#1}, \ref{#2}, \ref{#3} and \ref{#4}}
\def\secref#1{section~\ref{#1}}
\def\Secref#1{Section~\ref{#1}}
\def\twosecrefs#1#2{sections \ref{#1} and \ref{#2}}
\def\secrefs#1#2#3{sections \ref{#1}, \ref{#2} and \ref{#3}}
\def\eqref#1{equation~\ref{#1}}
\def\Eqref#1{Equation~\ref{#1}}
\def\plaineqref#1{\ref{#1}}
\def\chapref#1{chapter~\ref{#1}}
\def\Chapref#1{Chapter~\ref{#1}}
\def\rangechapref#1#2{chapters\ref{#1}--\ref{#2}}
\def\algref#1{algorithm~\ref{#1}}
\def\Algref#1{Algorithm~\ref{#1}}
\def\twoalgref#1#2{algorithms \ref{#1} and \ref{#2}}
\def\Twoalgref#1#2{Algorithms \ref{#1} and \ref{#2}}
\def\partref#1{part~\ref{#1}}
\def\Partref#1{Part~\ref{#1}}
\def\twopartref#1#2{parts \ref{#1} and \ref{#2}}

\def\ceil#1{\lceil #1 \rceil}
\def\floor#1{\lfloor #1 \rfloor}
\def\1{\bm{1}}
\newcommand{\train}{\mathcal{D}}
\newcommand{\valid}{\mathcal{D_{\mathrm{valid}}}}
\newcommand{\test}{\mathcal{D_{\mathrm{test}}}}

\def\eps{{\epsilon}}

\def\reta{{\textnormal{$\eta$}}}
\def\ra{{\textnormal{a}}}
\def\rb{{\textnormal{b}}}
\def\rc{{\textnormal{c}}}
\def\rd{{\textnormal{d}}}
\def\re{{\textnormal{e}}}
\def\rf{{\textnormal{f}}}
\def\rg{{\textnormal{g}}}
\def\rh{{\textnormal{h}}}
\def\ri{{\textnormal{i}}}
\def\rj{{\textnormal{j}}}
\def\rk{{\textnormal{k}}}
\def\rl{{\textnormal{l}}}
\def\rn{{\textnormal{n}}}
\def\ro{{\textnormal{o}}}
\def\rp{{\textnormal{p}}}
\def\rq{{\textnormal{q}}}
\def\rr{{\textnormal{r}}}
\def\rs{{\textnormal{s}}}
\def\rt{{\textnormal{t}}}
\def\ru{{\textnormal{u}}}
\def\rv{{\textnormal{v}}}
\def\rw{{\textnormal{w}}}
\def\rx{{\textnormal{x}}}
\def\ry{{\textnormal{y}}}
\def\rz{{\textnormal{z}}}

\def\rvepsilon{{\mathbf{\epsilon}}}
\def\rvtheta{{\mathbf{\theta}}}
\def\rva{{\mathbf{a}}}
\def\rvb{{\mathbf{b}}}
\def\rvc{{\mathbf{c}}}
\def\rvd{{\mathbf{d}}}
\def\rve{{\mathbf{e}}}
\def\rvf{{\mathbf{f}}}
\def\rvg{{\mathbf{g}}}
\def\rvh{{\mathbf{h}}}
\def\rvu{{\mathbf{i}}}
\def\rvj{{\mathbf{j}}}
\def\rvk{{\mathbf{k}}}
\def\rvl{{\mathbf{l}}}
\def\rvm{{\mathbf{m}}}
\def\rvn{{\mathbf{n}}}
\def\rvo{{\mathbf{o}}}
\def\rvp{{\mathbf{p}}}
\def\rvq{{\mathbf{q}}}
\def\rvr{{\mathbf{r}}}
\def\rvs{{\mathbf{s}}}
\def\rvt{{\mathbf{t}}}
\def\rvu{{\mathbf{u}}}
\def\rvv{{\mathbf{v}}}
\def\rvw{{\mathbf{w}}}
\def\rvx{{\mathbf{x}}}
\def\rvy{{\mathbf{y}}}
\def\rvz{{\mathbf{z}}}

\def\erva{{\textnormal{a}}}
\def\ervb{{\textnormal{b}}}
\def\ervc{{\textnormal{c}}}
\def\ervd{{\textnormal{d}}}
\def\erve{{\textnormal{e}}}
\def\ervf{{\textnormal{f}}}
\def\ervg{{\textnormal{g}}}
\def\ervh{{\textnormal{h}}}
\def\ervi{{\textnormal{i}}}
\def\ervj{{\textnormal{j}}}
\def\ervk{{\textnormal{k}}}
\def\ervl{{\textnormal{l}}}
\def\ervm{{\textnormal{m}}}
\def\ervn{{\textnormal{n}}}
\def\ervo{{\textnormal{o}}}
\def\ervp{{\textnormal{p}}}
\def\ervq{{\textnormal{q}}}
\def\ervr{{\textnormal{r}}}
\def\ervs{{\textnormal{s}}}
\def\ervt{{\textnormal{t}}}
\def\ervu{{\textnormal{u}}}
\def\ervv{{\textnormal{v}}}
\def\ervw{{\textnormal{w}}}
\def\ervx{{\textnormal{x}}}
\def\ervy{{\textnormal{y}}}
\def\ervz{{\textnormal{z}}}

\def\rmA{{\mathbf{A}}}
\def\rmB{{\mathbf{B}}}
\def\rmC{{\mathbf{C}}}
\def\rmD{{\mathbf{D}}}
\def\rmE{{\mathbf{E}}}
\def\rmF{{\mathbf{F}}}
\def\rmG{{\mathbf{G}}}
\def\rmH{{\mathbf{H}}}
\def\rmI{{\mathbf{I}}}
\def\rmJ{{\mathbf{J}}}
\def\rmK{{\mathbf{K}}}
\def\rmL{{\mathbf{L}}}
\def\rmM{{\mathbf{M}}}
\def\rmN{{\mathbf{N}}}
\def\rmO{{\mathbf{O}}}
\def\rmP{{\mathbf{P}}}
\def\rmQ{{\mathbf{Q}}}
\def\rmR{{\mathbf{R}}}
\def\rmS{{\mathbf{S}}}
\def\rmT{{\mathbf{T}}}
\def\rmU{{\mathbf{U}}}
\def\rmV{{\mathbf{V}}}
\def\rmW{{\mathbf{W}}}
\def\rmX{{\mathbf{X}}}
\def\rmY{{\mathbf{Y}}}
\def\rmZ{{\mathbf{Z}}}

\def\ermA{{\textnormal{A}}}
\def\ermB{{\textnormal{B}}}
\def\ermC{{\textnormal{C}}}
\def\ermD{{\textnormal{D}}}
\def\ermE{{\textnormal{E}}}
\def\ermF{{\textnormal{F}}}
\def\ermG{{\textnormal{G}}}
\def\ermH{{\textnormal{H}}}
\def\ermI{{\textnormal{I}}}
\def\ermJ{{\textnormal{J}}}
\def\ermK{{\textnormal{K}}}
\def\ermL{{\textnormal{L}}}
\def\ermM{{\textnormal{M}}}
\def\ermN{{\textnormal{N}}}
\def\ermO{{\textnormal{O}}}
\def\ermP{{\textnormal{P}}}
\def\ermQ{{\textnormal{Q}}}
\def\ermR{{\textnormal{R}}}
\def\ermS{{\textnormal{S}}}
\def\ermT{{\textnormal{T}}}
\def\ermU{{\textnormal{U}}}
\def\ermV{{\textnormal{V}}}
\def\ermW{{\textnormal{W}}}
\def\ermX{{\textnormal{X}}}
\def\ermY{{\textnormal{Y}}}
\def\ermZ{{\textnormal{Z}}}

\def\vzero{{\bm{0}}}
\def\vone{{\bm{1}}}
\def\vmu{{\bm{\mu}}}
\def\vtheta{{\bm{\theta}}}
\def\vphi{{\bm{\phi}}}
\def\vsigma{{\bm{\sigma}}}
\def\vomega{{\bm{\omega}}}
\def\va{{\bm{a}}}
\def\vb{{\bm{b}}}
\def\vc{{\bm{c}}}
\def\vd{{\bm{d}}}
\def\ve{{\bm{e}}}
\def\vf{{\bm{f}}}
\def\vg{{\bm{g}}}
\def\vh{{\bm{h}}}
\def\vi{{\bm{i}}}
\def\vj{{\bm{j}}}
\def\vk{{\bm{k}}}
\def\vl{{\bm{l}}}
\def\vm{{\bm{m}}}
\def\vn{{\bm{n}}}
\def\vo{{\bm{o}}}
\def\vp{{\bm{p}}}
\def\vq{{\bm{q}}}
\def\vr{{\bm{r}}}
\def\vs{{\bm{s}}}
\def\vt{{\bm{t}}}
\def\vu{{\bm{u}}}
\def\vv{{\bm{v}}}
\def\vw{{\bm{w}}}
\def\vx{{\bm{x}}}
\def\vy{{\bm{y}}}
\def\vz{{\bm{z}}}

\def\evalpha{{\alpha}}
\def\evbeta{{\beta}}
\def\evepsilon{{\epsilon}}
\def\evlambda{{\lambda}}
\def\evomega{{\omega}}
\def\evmu{{\mu}}
\def\evpsi{{\psi}}
\def\evsigma{{\sigma}}
\def\evtheta{{\theta}}
\def\eva{{a}}
\def\evb{{b}}
\def\evc{{c}}
\def\evd{{d}}
\def\eve{{e}}
\def\evf{{f}}
\def\evg{{g}}
\def\evh{{h}}
\def\evi{{i}}
\def\evj{{j}}
\def\evk{{k}}
\def\evl{{l}}
\def\evm{{m}}
\def\evn{{n}}
\def\evo{{o}}
\def\evp{{p}}
\def\evq{{q}}
\def\evr{{r}}
\def\evs{{s}}
\def\evt{{t}}
\def\evu{{u}}
\def\evv{{v}}
\def\evw{{w}}
\def\evx{{x}}
\def\evy{{y}}
\def\evz{{z}}

\def\mA{{\bm{A}}}
\def\mB{{\bm{B}}}
\def\mC{{\bm{C}}}
\def\mD{{\bm{D}}}
\def\mE{{\bm{E}}}
\def\mF{{\bm{F}}}
\def\mG{{\bm{G}}}
\def\mH{{\bm{H}}}
\def\mI{{\bm{I}}}
\def\mJ{{\bm{J}}}
\def\mK{{\bm{K}}}
\def\mL{{\bm{L}}}
\def\mM{{\bm{M}}}
\def\mN{{\bm{N}}}
\def\mO{{\bm{O}}}
\def\mP{{\bm{P}}}
\def\mQ{{\bm{Q}}}
\def\mR{{\bm{R}}}
\def\mS{{\bm{S}}}
\def\mT{{\bm{T}}}
\def\mU{{\bm{U}}}
\def\mV{{\bm{V}}}
\def\mW{{\bm{W}}}
\def\mX{{\bm{X}}}
\def\mY{{\bm{Y}}}
\def\mZ{{\bm{Z}}}
\def\mBeta{{\bm{\beta}}}
\def\mPhi{{\bm{\Phi}}}
\def\mLambda{{\bm{\Lambda}}}
\def\mSigma{{\bm{\Sigma}}}
\def\mTheta{{\bm{\Theta}}}
\def\mZero{{\bm{0}}}

\newcommand{\tens}[1]{\bm{\mathsfit{#1}}}
\def\tA{{\tens{A}}}
\def\tB{{\tens{B}}}
\def\tC{{\tens{C}}}
\def\tD{{\tens{D}}}
\def\tE{{\tens{E}}}
\def\tF{{\tens{F}}}
\def\tG{{\tens{G}}}
\def\tH{{\tens{H}}}
\def\tI{{\tens{I}}}
\def\tJ{{\tens{J}}}
\def\tK{{\tens{K}}}
\def\tL{{\tens{L}}}
\def\tM{{\tens{M}}}
\def\tN{{\tens{N}}}
\def\tO{{\tens{O}}}
\def\tP{{\tens{P}}}
\def\tQ{{\tens{Q}}}
\def\tR{{\tens{R}}}
\def\tS{{\tens{S}}}
\def\tT{{\tens{T}}}
\def\tU{{\tens{U}}}
\def\tV{{\tens{V}}}
\def\tW{{\tens{W}}}
\def\tX{{\tens{X}}}
\def\tY{{\tens{Y}}}
\def\tZ{{\tens{Z}}}

\def\gA{{\mathcal{A}}}
\def\gB{{\mathcal{B}}}
\def\gC{{\mathcal{C}}}
\def\gD{{\mathcal{D}}}
\def\gE{{\mathcal{E}}}
\def\gF{{\mathcal{F}}}
\def\gG{{\mathcal{G}}}
\def\gH{{\mathcal{H}}}
\def\gI{{\mathcal{I}}}
\def\gJ{{\mathcal{J}}}
\def\gK{{\mathcal{K}}}
\def\gL{{\mathcal{L}}}
\def\gM{{\mathcal{M}}}
\def\gN{{\mathcal{N}}}
\def\gO{{\mathcal{O}}}
\def\gP{{\mathcal{P}}}
\def\gQ{{\mathcal{Q}}}
\def\gR{{\mathcal{R}}}
\def\gS{{\mathcal{S}}}
\def\gT{{\mathcal{T}}}
\def\gU{{\mathcal{U}}}
\def\gV{{\mathcal{V}}}
\def\gW{{\mathcal{W}}}
\def\gX{{\mathcal{X}}}
\def\gY{{\mathcal{Y}}}
\def\gZ{{\mathcal{Z}}}

\def\sA{{\mathbb{A}}}
\def\sB{{\mathbb{B}}}
\def\sC{{\mathbb{C}}}
\def\sD{{\mathbb{D}}}
\def\sF{{\mathbb{F}}}
\def\sG{{\mathbb{G}}}
\def\sH{{\mathbb{H}}}
\def\sI{{\mathbb{I}}}
\def\sJ{{\mathbb{J}}}
\def\sK{{\mathbb{K}}}
\def\sL{{\mathbb{L}}}
\def\sM{{\mathbb{M}}}
\def\sN{{\mathbb{N}}}
\def\sO{{\mathbb{O}}}
\def\sP{{\mathbb{P}}}
\def\sQ{{\mathbb{Q}}}
\def\sR{{\mathbb{R}}}
\def\sS{{\mathbb{S}}}
\def\sT{{\mathbb{T}}}
\def\sU{{\mathbb{U}}}
\def\sV{{\mathbb{V}}}
\def\sW{{\mathbb{W}}}
\def\sX{{\mathbb{X}}}
\def\sY{{\mathbb{Y}}}
\def\sZ{{\mathbb{Z}}}

\def\emLambda{{\Lambda}}
\def\emA{{A}}
\def\emB{{B}}
\def\emC{{C}}
\def\emD{{D}}
\def\emE{{E}}
\def\emF{{F}}
\def\emG{{G}}
\def\emH{{H}}
\def\emI{{I}}
\def\emJ{{J}}
\def\emK{{K}}
\def\emL{{L}}
\def\emM{{M}}
\def\emN{{N}}
\def\emO{{O}}
\def\emP{{P}}
\def\emQ{{Q}}
\def\emR{{R}}
\def\emS{{S}}
\def\emT{{T}}
\def\emU{{U}}
\def\emV{{V}}
\def\emW{{W}}
\def\emX{{X}}
\def\emY{{Y}}
\def\emZ{{Z}}
\def\emSigma{{\Sigma}}

\newcommand{\etens}[1]{\mathsfit{#1}}
\def\etLambda{{\etens{\Lambda}}}
\def\etA{{\etens{A}}}
\def\etB{{\etens{B}}}
\def\etC{{\etens{C}}}
\def\etD{{\etens{D}}}
\def\etE{{\etens{E}}}
\def\etF{{\etens{F}}}
\def\etG{{\etens{G}}}
\def\etH{{\etens{H}}}
\def\etI{{\etens{I}}}
\def\etJ{{\etens{J}}}
\def\etK{{\etens{K}}}
\def\etL{{\etens{L}}}
\def\etM{{\etens{M}}}
\def\etN{{\etens{N}}}
\def\etO{{\etens{O}}}
\def\etP{{\etens{P}}}
\def\etQ{{\etens{Q}}}
\def\etR{{\etens{R}}}
\def\etS{{\etens{S}}}
\def\etT{{\etens{T}}}
\def\etU{{\etens{U}}}
\def\etV{{\etens{V}}}
\def\etW{{\etens{W}}}
\def\etX{{\etens{X}}}
\def\etY{{\etens{Y}}}
\def\etZ{{\etens{Z}}}

\newcommand{\pdata}{p_{\rm{data}}}
\newcommand{\ptrain}{\hat{p}_{\rm{data}}}
\newcommand{\Ptrain}{\hat{P}_{\rm{data}}}
\newcommand{\pmodel}{p_{\rm{model}}}
\newcommand{\Pmodel}{P_{\rm{model}}}
\newcommand{\ptildemodel}{\tilde{p}_{\rm{model}}}
\newcommand{\pencode}{p_{\rm{encoder}}}
\newcommand{\pdecode}{p_{\rm{decoder}}}
\newcommand{\precons}{p_{\rm{reconstruct}}}

\newcommand{\laplace}{\mathrm{Laplace}} 

\newcommand{\E}{\mathbb{E}}
\newcommand{\Ls}{\mathcal{L}}
\newcommand{\R}{\mathbb{R}}
\newcommand{\emp}{\tilde{p}}
\newcommand{\lr}{\alpha}
\newcommand{\reg}{\lambda}
\newcommand{\rect}{\mathrm{rectifier}}
\newcommand{\softmax}{\mathrm{softmax}}
\newcommand{\sigmoid}{\sigma}
\newcommand{\softplus}{\zeta}
\newcommand{\KL}{D_{\mathrm{KL}}}
\newcommand{\Var}{\mathrm{Var}}
\newcommand{\standarderror}{\mathrm{SE}}
\newcommand{\Cov}{\mathrm{Cov}}
\newcommand{\normlzero}{L^0}
\newcommand{\normlone}{L^1}
\newcommand{\normltwo}{L^2}
\newcommand{\normlp}{L^p}
\newcommand{\normmax}{L^\infty}

\newcommand{\parents}{Pa} 

\let\ab\allowbreak

%% file: data/introduction.tex
\vspace{-0.2cm}
\section{Introduction}

Facial expression (FE) is one of the most powerful, natural, and universal symbols for human beings to convey their emotions. In various human-computer interaction applications, automatic facial expression recognition (FER) can provide more insights into users' emotional states and intentions, which is essential for achieving better behavioral responses in driver fatigue surveillance, sociable robotics, and healthcare. To improve the robustness of FER in these real-world applications, tremendous efforts have been made from different perspectives, such as occluded facial image [34], noisy labels [25, 29], cross-dataset generalization [15], and so on [30, 44]. However, FER under extreme lighting conditions is still an open question. 

\begin{figure}[t]
  \centering\includegraphics[width=0.75\linewidth]{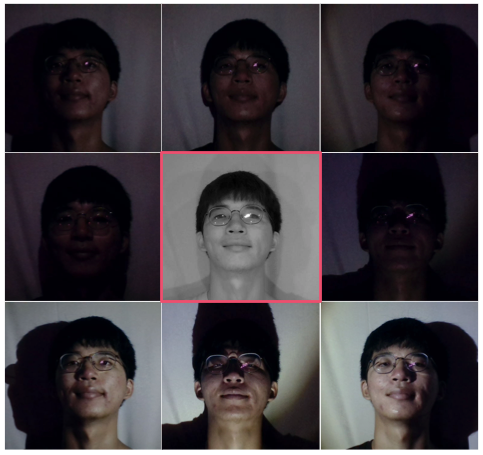}
  \vspace{-2mm}
   \caption{Comparison of VIS and NIR facial images under extreme lighting. The red frame in the center of the figure is NIR images. Even under low lighting or complete darkness, the geometric and texture details of volunteers in NIR images are still very well preserved.}
   \label{fig1}
  \vspace{-4mm}
  \end{figure}

As illustrated in Fig 1, it’s quite difficult for visible (VIS) sensors to capture high-quality images in low lighting or complete darkness conditions. The limitations of these sensors result in the loss of crucial geometric and textural details, especially in key areas such as the cheek or lips. Consequently, even experienced annotators find it arduous to accurately identify the expression type depicted in such images. Hence, it becomes exceedingly challenging to solely rely on algorithmic enhancements to address this issue. Fortunately, the implementation of near-infrared (NIR) imaging techniques offers a low-cost and effective solution to enhance the performance of VIS FER systems in low light conditions. Furthermore, given their reduced sensitivity to visible (VIS) light illumination variations, NIR face sensors have been widely used in security surveillance within finance, transportation, and other related domains.

However, achieving better performance for NIR FER is still a challenging problem for deep models and remains largely unresolved due to the following two reasons: 1) Insufficient training samples. In the field of VIS, after decades of hard work, more and more high-quality data sets, such as CK+, RAF-DB, KDEF, etc., continue to emerge, and the overall amount of available training data is also growing rapidly. In contrast, there are currently only two NIR facial expression datasets available, Oulu-CASIA (80 subjects) and Large-HFE (360 subjects). 2) The large modal discrepancy. VIS and NIR face images taken of the same individual are obtained through distinct sensory devices with varying settings, resulting in significant differences in their visual appearances. Thus, directly applying deep models trained on VIS data fails to provide satisfactory results in the NIR spectrum. 

To tackle these challenges together, synthesis methods pose opportunities to can generate NIR images given input VIS face images, so that abundant label information will migrate from VIS to NIR. However, existing methods usually synthesize images without prior task knowledge. Thus, they can not realize customizable domain transfer that allows a controllable appearance for FER. According to \cite{zhang2011facial}, the feature of facial expression usually lies in subtle facial movements, which requires detailed facial representations to be preserved in the facial expression translation process.

Inspired by these observations, we present a novel method, called NFER-SDE, to learn and synthesize NIR expression images of good quality from VIS. The synthesized facial images not only preserve the high-frequency content carrying modality-invariant information, but also contains detailed facial representations which is crucial to FER tasks. Specifically, NFER-SDE consists of two novel components, energy functions of task-specific guidance and conditional score matching. First, a novel energy function $\gE_h$ is proposed to guide the reverse SDE process using facial landmark information. Thus, the key facial structure information can be preserved through the reverse SDE process, resulting in lower uncertainty. Encouraged by unique observation in the cross-modality facial expression translation task, the frequency sharing rule is established to preserve more modality-invariant information. Second, conditional score matching is introduced to make sure that the generated NIR image can be more faithful to the original VIS image. 

In summary, our main contributions are three-fold:

(1)	We develop a novel method, i.e. NFER-SDE, to translate the VIS facial expression sample to NIR modality. To the best of our knowledge, this is the first effort to synthesize NIR facial expression samples for the NIR FER task.

(2)	We propose new task-specific energy functions to guide the image generation process with the domain knowledge of FER. We also employ a conditional score-matching network to better focus on facial expression details from the original image.

(3) We conduct extensive experiments on Oulu-CASIA and Large-HFE. The results show that our method has a better performance than the state-of-art image synthesis approaches.

%% file: data/related_work.tex
\vspace{-0.2cm}
\section{Related Work}
\paragraph{Facial Expression Recognition} The goal of FER is to automatically encode emotional information from subtle facial muscle changes. Inspired by the success of deep neural network, extensive efforts have been made to address different problems including occluded facial image \cite{xing2022co}, noisy labels \cite{wang2022ease}, cross-dataset generalization \cite{li2020deeper}, dynamic FER \cite{zhao2021former}. As discussed in [3], dealing with the challenge of illumination variations is also crucial to FER in real-world applications. NIR images provide a low-cost and effective solution to enhance the performance of VIS FER systems in uncontrolled or non-visible light conditions. In this paper, we present the first approach that transforms face expression appearance between heterogeneous modalities for NIR FER in extreme light conditions.

\paragraph{Image-to-image Translation} Image-to-image translation is an image generation task to transform the images between two different visual domains.
In this field, various generative models have made great progress, including Generative adversarial networks (GAN) \cite{isola2017image, yang2020hifacegan}, Variational AutoEncoders (VAE) \cite{kingma2014semi, sohn2015learning, peng2021generating}.
Recently, diffusion models have attracted much attention due to their remarkable image generation ability in terms of distribution diversity \cite{croitoru2023diffusion}, and thus are utilized in the image-to-image translation task. ILVR \cite{choi2021ilvr} proposes an iterative latent variable refinement strategy to guide the generative process of diffusion models. EGSDE \cite{zhao2022egsde} employs a pretrained energy function to guide the reverse process of score-based diffusion models.

%% file: data/method.tex
\vspace{-0.2cm}
\section{Problem Definition}
Let $\gD_\text{HTR}=\{(\vx_V^{(i)}, \vx_N^{(i)}, y^{(i)})|i=1,\cdots, n\}$ denote an NIR-VIS heterogeneous facial expression dataset, where $\vx_V^{(i)}$ and $\vx_N^{(i)}$ are the VIS and NIR images of the $i$-th sample respectively, and $y^{(i)}$ is the corresponding expression label. Let $\gD_\text{VIS}=\{(\vx^{(j)}, y^{(j)})|j=1,\cdots, m\}$ denote a VIS facial expression dataset, where $\vx^{(j)}$ is the VIS image of the $j$-th sample and $y^{(j)}$ is the corresponding label. The goal of NIR facial expression synthesis is to learn a translation model $F$ from the heterogeneous dataset $\gD_\text{HTR}$, i.e., the model can transform arbitrary facial expression image $\vx_V$ of VIS modality to the NIR modality $\vx_I=F(\vx_V)$. 
Using this translation model, we can create a synthesized NIR-VIS facial expression dataset $\gD_\text{SYS}=\{(\vx^{(j)}, F(\vx^{(j)}), y^{(j)})| j=1,\cdots, j\}$ from the VIS dataset $\gD_\text{VIS}$, where $F(\vx^{(j)})$ is the synthesized NIR image of the $j$-th sample. 
The final goal for NIR facial expression image synthesis is that the FER backbone trained on the augmented dataset $\gD_\text{HTR}\cup\gD_\text{VIS}$ can achieve better performance than that only trained on $\gD_\text{HTR}$.

\vspace{-0.2cm}
\section{Preliminaries}
\subsection{Diffusion Models}
Diffusion models are a class of deep generative models, which have achieved remarkable performance in image generation over previous state-of-the-art methods like GANs \cite{goodfellow2014generative}.
The core idea of diffusion models is to gradually corrupt the training data structure by adding noise to the input, and then learn a reverse network to recover the original data from the corrupted data \cite{croitoru2023diffusion}.

The training process of diffusion models is formulated as two stages: the forward stage and the reverse stage. In the forward stage, random Gaussian noise is iteratively added to the input, with the proportion of noise increasing from $0$ to $1$. In other words, the input is gradually degraded into pure noise. The reverse stage is a noise-removal process, which is the reversal of the preceding forward noise-adding process. In the reverse stage, the output of the forward stage is recovered iteratively using a denoising neural network that is trained to predict the added noise at each step. 

During sampling, random noise is sampled from the Gaussian distribution and then directly fed into the reverse stage to generate the corresponding sample. Therefore, diffusion models are able to learn the distribution of training data and generate samples from random noise by the reverse network.
\vspace{-0.2cm}
\subsection{Score-based Diffusion Models (SBDMs)}
Score-Based Diffusion Models (SBDMs) are a continuous form of the generic diffusion model. By transforming the above discrete diffusion process into a continuous form, SBDMs utilize Stochastic Differential Equation (SDE) to model and solve the diffusion process \cite{song2020score}. 

Let $\{\vx_t\}_{t=0}^T$ denote the image sequence in the forward stage, where $\vx_0$ is the original image and $\vx_T$ is the final degraded image. In SBDMs, the time index $t$ is regarded as a continuous variable $t\in[0,T]$, and the image sequence $\{\vx_t\}_{t=0}^T$ is generalized into a continuous function $\vx_t=\vx(t)$. Thus, the forward stage can be formulated as the corresponding SDE
\begin{equation}\label{eq:forward-sde}
    d\vx_t = \vf(\vx_t, t)dt + \sigma(t)d\vomega
\end{equation}
where $\vf$ is the drift coefﬁcient of $
\vx(t)$ that depends on the image $\vx(t)$ and time $t$, $\sigma$ is the diffusion coefficient of $
\vx(t)$ that only depends on time $t$, and $\vomega$ is the integral of the white noise Gaussian process, i.e., the standard Wiener process.
According to \cite{song2020score}, general DDPM converges to the Variance Preserving SDE as the number of sample steps $T\rightarrow \infty$, where $\vf(\vx_t, t)=-\frac{1}{2}\beta(t)\vx(t)$, $\sigma(t)=\sqrt{\beta(t)}$. Thus, DDPM networks can be utilized to implement SBDMs.

Given the forward SDE in Eq. (\ref{eq:forward-sde}), the corresponding reverse SDE is solved as
\begin{equation}\label{eq:reverse-sde}
    d\vx_t = [\vf(\vx_t, t)-\sigma(t)^2\nabla_{\vx_t}\log p(\vx_t)]dt+\sigma(t)d\hat{\vomega}
\end{equation}
where $p(\vx_t)$ is the marginal data distribution at time $t$, and $\hat{\vomega}$ is the reverse Wiener process from $T$ to $0$. 
In order to estimate the marginal data distribution, SBDMs propose a score-matching neural network $\vs(\vx_t, t, \mTheta_s)$ to parameterize the score function $\nabla_{\vx_t}\log p(\vx_t)$, where $\mTheta_s$ is the parameter of the network. Using the Euler-Maruyama method, the reverse SDE can be further discretized as
\begin{equation}\label{eq:reverse-iter}
    \Delta \vx = [\vf(\vx_t, t)-\sigma(t)^2\vs(\vx_t, t, \mTheta_s)]\Delta t+\sigma(t)\sqrt{|\Delta t|} \vz
\end{equation}
where $\Delta t < 0$ is the reverse step size, and $\vz\sim \gN(\mZero, \mI)$ is a standard Gaussian noise. Eq. (\ref{eq:reverse-iter}) provides the iterative rule for the reverse stage, which enables the computational simulation of SDE for unconditional image generation.
\vspace{-0.2cm}
\subsection{Energy-Guided Conditional SBDMs}
In order to generate images under specific requirements or conditions, the conditional form of SBDMs is proposed to incorporate the conditional information into the diffusion process \cite{song2020score}. Given a condition $\vc$, the data distribution of $\vx_t$ turns into the conditional distribution $p(\vx_t|\vc)$. According to Bayesian rule, the score function of the target data distribution is formulated as $\nabla_{\vx_t}\log p(\vx_t|\vc)=\nabla_{\vx_t}\log p(\vx_t) + \nabla_{\vx_t} \log p(\vc|\vx_t)$. Thus, the reverse SDE in Eq. (\ref{eq:reverse-sde}) is modified as
\begin{equation}
    \begin{aligned}
        d\vx_t = &[\vf(\vx_t, t)-\sigma(t)^2(\nabla_{\vx_t}\log p(\vx_t)\\
        &+\nabla_{\vx_t}\log p(\vc|\vx_t))]dt+\sigma(t)d\hat{\vomega}
    \end{aligned}
\end{equation}
where $\nabla_{\vx_t}\log p(\vc|\vx_t)$ is added to the original score function as a modification term. Given the condition representation model $p(\vc|\vx_t)$, the reverse SDE can be calculated under the guidance of the corresponding condition. The representation model $p(\vc|\vx_t)$ is exactly matched with some special cases like classifier guidance \cite{zhao2022egsde}. However, in some more cases like image translation, the conditional distribution $p(\vc|\vx_t)$ is intractable.

In order to obtain a more generalized form of guidance, the conditional SBDMs further propose an energy function $\gE(\vx_t, \vc, t)$ to parameterize the modification term $\nabla_{\vx_t}\log p(\vc|\vx_t)$ \cite{zhao2022egsde}. The energy function $\gE(\vx_t, \vc, t)$ is trained or designed to be minimized when the condition is satisfied. Thus, the modiﬁcation term is formulated as $\nabla_{\vx_t}\log p(\vc|\vx_t)=-\nabla_{\vx_t}\gE(\vx_t, \vc, t)$. Given the score matching network $\vs(\vx_t, t, \mTheta_s)$ and the energy function $\gE(\vx_t, \vc, t)$, the reverse SDE is formulated as
\begin{equation}\label{eq:energy-guided-reverse-sde}
    \begin{aligned}
        d\vx_t = &[\vf(\vx_t, t)-\sigma(t)^2(\vs(\vx_t, t, \mTheta_s)\\
        &-\nabla_{\vx_t}\gE(\vx_t, \vc, t))]dt+\sigma(t)d\hat{\vomega}
    \end{aligned}
\end{equation}
With the guidance of different energy functions, Eq. (\ref{eq:energy-guided-reverse-sde}) can be applied to various conditional image generation tasks.
\vspace{-0.2cm}
\section{Method}
In this section, we present the proposed method for VIS-NIR facial expression image translation. Firstly, we introduce the framework of the proposed method. Then, we describe the energy function of task-specific guidance and the conditional score matching network in detail.

\begin{figure*}[htbp]
    \centering
    \vspace{-0.2cm}
    \includegraphics[width=0.99\textwidth]{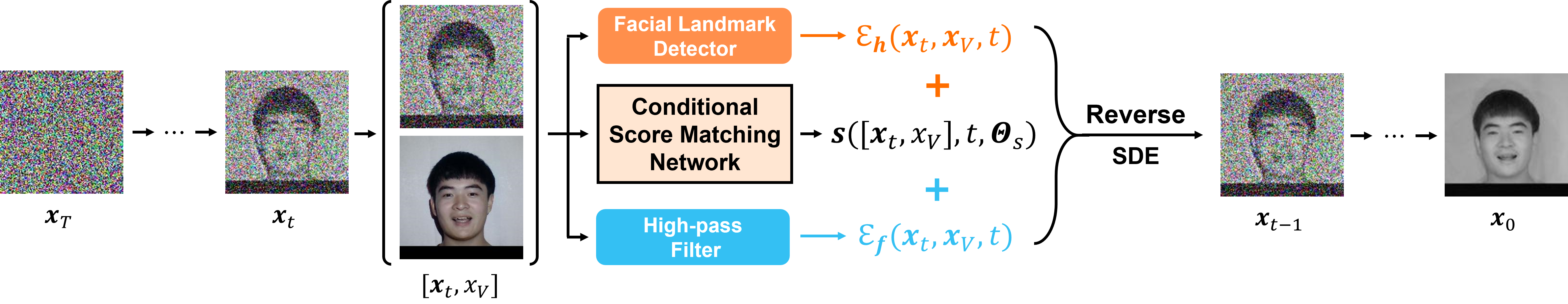}
    \caption{The main framework of the proposed NFER-SDE network for VIS-NIR facial expression translation. The framework mainly consists of two task-specific guidance energy functions and a conditional score-matching network.}
    \label{fig:framework}
    \vspace{-0.2cm}
\end{figure*}

\vspace{-0.2cm}
\subsection{Model Overview}
The framework of the proposed method is illustrated in Fig. \ref{fig:framework}. Specifically, the proposed NFER-SDE consists of two main components. The first component is the newly designed energy functions of task-specific guidance, which can guide the image-generation process with the domain-specific knowledge of facial expression, including information on the key facial areas and the unique frequency-sharing rule of cross-modality expression translation.
The second component is the conditional score-matching network, which can make full use of the subtle facial details in the original facial expression image by explicitly conditioning it as part of the input of the score-matching network.
\vspace{-0.2cm}
\subsection{Energy Functions of Task-Specific Guidance}
Conditional diffusion models \cite{choi2021ilvr, meng2021sdedit, zhao2022egsde} have made great progress and achieved significant results in general image-to-image translation tasks, such as image editing and style transfer. 
Yet, when applied to VIS-NIR facial expression translation task, the existing methods do not perform as expected in transferring facial expression attributes. 
The underperformance is partly due to the feature learning challenges in VIS-NIR facial expression translation task \cite{wang2020facial}, which can be summarized in the following two aspects.

Firstly, compared to general vision targets, facial expressions are usually conveyed through more subtle facial representations in specific facial areas \cite{zhu2023knowledge}. This requires the diffusion models to be well guided by key facial structure information that is related to facial expression. 
Recently, it is widely recognized that the facial landmark coordinates can encode the facial structure information, which is crucial for facial expression recognition \cite{lv20193d, wang2020suppressing}. 
Therefore, a novel energy function $\gE_h$ is proposed to guide the reverse SDE process using facial landmark information.
Specifically, a time-dependent facial landmark detection network $\mH(\vx, t, \mTheta_h)$ is proposed to extract the facial landmark heatmap from the input image $\vx$ and current step $t$, with the parameter $\mTheta_h$. Let $\vc_t$ denote the perturbed source image in the forward SDE and $\vx_t$ denote the denoised target image in the reverse SDE both at step $t$. The energy function is formulated as
\begin{equation}
    \gE_h(\vx_t, \vc, t) = \frac{1}{2}\|\mH(\vx_t, t, \mTheta_h)-\mH(\vc_t, t, \mTheta_h)\|_2^2
    \label{eq:landmark-energy}
\end{equation}
This energy function is designed to be minimized when the facial landmark heatmap of the target image $\vx_t$ is close to that of the perturbed source image $\vc_t$. Thus, the key facial structure information can be preserved through the reverse SDE process.

\begin{figure}[t]
    \centering
    \vspace{-0.2cm}
    \includegraphics[width=0.8\columnwidth]{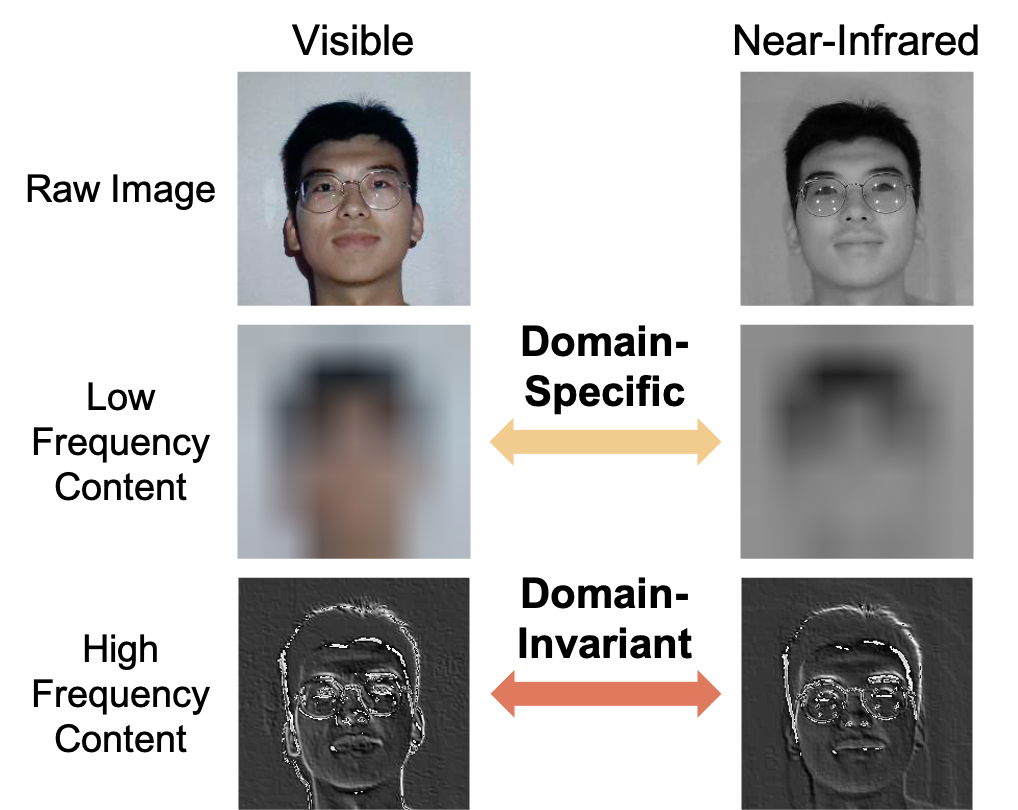} 
    \caption{Visualization of frequency sharing rule in cross-modality facial expression translation. It can be observed that the high-frequency content carries more modality-invariant information that should be preserved in cross-modality facial expression translation.}
    \label{fig:frequency}
    \vspace{-0.2cm}
\end{figure}

Secondly, the frequency sharing rule in cross-modality translation task is quite unique, as shown in Fig. \ref{fig:frequency}. In the cross-modality facial expression translation task, the high-frequency content of the image carries more modality-invariant information for facial expression translation, as it contains more detailed facial expression representations. 
This rule is quite different from that in general image translation tasks where the low-frequency content is regarded to carry the domain-invariant information that should be shared between the source and target domains \cite{choi2021ilvr,zhao2022egsde}. 
Thus, the low-pass filters that general image translation tasks adopt to preserve low-frequency content are not suitable for the cross-modality facial expression translation task any more.

Therefore, we propose a novel energy function based on high-pass filter to preserve the high-frequency content in the reverse SDE process. 
Given a high-pass filter $\Phi(\cdot)$ as the domain-invariant feature extractor, the energy function is formulated as
\begin{equation}
    \gE_f(\vx_t, \vc, t) = \|\Phi(\vx_t)-\Phi(\vc_t)\|_2^2
    \label{eq:high-frequency-energy}
\end{equation}
With the guidance of $\gE_f$, the high-frequency content can be preserved through the reverse SDE process to preserve the domain-invariant feature in the cross-modality translation task.

To summarize, the above two energy functions are aggregated to form the final energy function
\begin{equation}
    \gE(\vx_t, \vc, t) = \lambda_h\gE_h(\vx_t, \vc, t) + \lambda_f\gE_f(\vx_t, \vc, t)
\end{equation}
where $\lambda_h$ and $\lambda_f$ are the weighting hyper-parameters respectively.

\vspace{-0.2cm}
\subsection{Conditional Score Matching}
According to \cite{zhang2011facial}, the feature of facial expression usually lies in subtle facial movements, which requires detailed facial representations to be preserved in the facial expression translation process. Thus, in the VIS-NIR facial expression translation task, the original VIS image should be fully considered by the denoising network $\vs(\vx_t, t, \mTheta_s)$ to ensure faithfulness of the generated NIR image.
However, existing SDE-based models \cite{meng2021sdedit,choi2021ilvr,zhao2022egsde} do not directly take the original image information into the denoising neural network at each iterative reverse step, which may cause the generated image to lose some key facial details.

To solve the above problem, we propose to condition the denoising neural network with the reference image $\vc$ in the cross-modality translation process. 
Inspired by \cite{saharia2022image}, the reference image $\vc$ is concatenated to the input image $\vx_t$ in the channel dimension at each step $t$. The concatenated vector is then fed into the score matching network to generate the output 
\begin{equation}
    \vs_t = \vs([\vx_t, \vc], t, \mTheta_s)
\end{equation}
where $\vs(\cdot, \cdot, \mTheta_s)$ is the extension of that in Eq. (\ref{eq:energy-guided-reverse-sde}). The training algorithm of the proposed conditional score matching network is summarized in Algorithm \ref{alg:algorithm-train}.

\begin{algorithm}[htbp]
\caption{Training conditional score matching network}
\label{alg:algorithm-train}
\textbf{Require}: The training dataset $\{(\vx_V^{(i)}, \vx_N^{(i)}, y^{(i)})\}_{i=1}^n$, the conditional score matching network $\vs(\cdot, \cdot, \mTheta_s)$, the number of steps $T$.
\begin{algorithmic}[1] 
\REPEAT
\STATE Take a sample pair $(\vx_V, \vx_N)$ from the training dataset $\{(\vx_V^{(i)}, \vx_N^{(i)}, y^{(i)})\}_{i=1}^n$.
\STATE Take a step $t \sim \gU\{1, T\}$.
\STATE Take a random noise $\rvepsilon \sim \gN(\mZero, \rmI)$.
\STATE $\vx_t=\sqrt{\overline{\alpha}_t}\vx_N + \sqrt{1-\overline{\alpha}_t}\rvepsilon$
\STATE Stochastic gradient descent step on gradient $\nabla_{\mTheta_s} \Vert \vs([\vx_t, \vx_V], t, \mTheta_s)-\frac{-\rvepsilon}{\sqrt{1-\overline{\alpha}_t}}\Vert_2^2$
\UNTIL converged
\end{algorithmic}
\end{algorithm}

Using the additional input $\vc$ during training, the new conditioned denoising network is supervised to generate better NIR images under the guidance of the detailed information from the VIS source image. Thus, the generated NIR image can be more faithful to the original VIS image.

%% file: data/experiment.tex
\vspace{-0.2cm}
\section{Experiment Setup}
\subsection{Evaluation Settings}
In order to obtain comprehensive evaluation of NFER-SDE, we follow previous works in heterogeneous translation \cite{yang2020hifacegan,nair2023t2v} and design two settings as follows: improved NIR FER performance and VIS-NIR translation quality.
For \textbf{improved NIR FER performance,} extra VIS facial expression samples are translated to NIR modality, and then added to the NIR training samples of original NIR\&VIS facial expression dataset. The augmented NIR training samples are utilized to train FER benchmarks. The well-trained FER benchmarks are then evaluated on the original NIR test set and the accuracy and macro F1 score are reported as the improved NIR FER performance.
For \textbf{VIS-NIR translation quality}, the NIR images generated from the VIS test samples of the original NIR\&VIS facial expression dataset are compared with the ground-truth NIR test samples of the original dataset according to the image quality metrics including Learned Perceptual Image Patch Similarity (LPIPS) \cite{zhang2018unreasonable}, Peak Signal-to-Noise Ratio (PSNR), Structural Similarity (SSIM).

\vspace{-0.2cm}
\subsection{Datasets}
For NIR\&VIS facial expression datasets, we select the only two available ones: Oulu-CASIA \cite{zhao2011facial} and Large-HFE. \textbf{Oulu-CASIA} is a laboratory-controlled dataset that consists of video sequences captured with both visible and near-infrared lights from 80 subjects. \textbf{Large-HFE} is also captured in the laboratory environment, with a larger size of 360 subjects. In both datasets, only the six basic expressions are considered. For each dataset, we employ the subject-independent strategy to split it into a train set of $80\%$ subjects and a test set of $20\%$ subjects. The train set is utilized to train the proposed model and the test set is used to evaluate the performance.

For extra VIS facial expression dataset, we select CFEE \cite{du2014compound} dataset. \textbf{CFEE} captures facial expression images in the laboratory with visible light from 230 subjects, with neutral and 6 basic expressions and 15 compound expressions. Only the 6 basic expressions are used in our experiments. The full set of CFEE is utilized to be transformed into NIR modality by the proposed model and the transformed samples are added to the training set of Oulu-CASIA or Large-HFE to jointly train the NIR FER benchmark.
\vspace{-0.2cm}
\subsection{Baselines}
In order to comprehensively evaluate the proposed method, We select the following image-to-image translation baselines for comparison, including the GAN-based methods and diffusion-based methods. For GAN-based methods, we select the state-of-the-art methods in image translation, including pix2pix \cite{isola2017image}, CycleGAN \cite{zhu2017unpaired}, HiFaceGAN \cite{yang2020hifacegan}. For diffusion-based methods, we select the state-of-the-art methods in image restoration, including two parts: the denoising diffusion probabilistic models (DDPMs) and score-based diffusion models (SBDMs). The DDPM baselines used in this paper include 
ILVR \cite{choi2021ilvr}, SR3 \cite{saharia2022image}, and T2V-DDPM \cite{nair2023t2v}. The SBDM baselines include SDEdit \cite{meng2021sdedit} and EGSDE \cite{zhao2022egsde}. Among these baselines, HiFaceGAN and T2V-DDPM are specially designed for heterogeneous face translation, while the others are general image translation methods. For fair comparison, we train all the baselines on the same training set and evaluate them on the same test set.

\begin{table*}[t]
    \centering
    \vspace{-5mm}
    \caption{The performance of the compared methods on Oulu-CASIA. \dag\  denotes methods specially designed for heterogeneous face translation. The best results are highlighted in \textbf{bold}.}
    \vspace{-3mm}
    \begin{tabular}{cl|cc|ccc}
        \toprule
        \multicolumn{2}{c|}{\multirow{2}[2]{*}{Method}} & \multicolumn{2}{c|}{Performance} & \multicolumn{3}{c}{Image Quality} \\
        \multicolumn{2}{c|}{} & Acc ($\%$) & F1 ($\%$) & LPIPS (↓) & PSNR (↑) & SSIM(↑) \\
        \midrule
        \multicolumn{2}{c|}{Without translation} & $61.11$ & $61.44$ & $0.6369$ & $48.45$ & $0.9069$ \\
        \midrule
        \multicolumn{1}{c}{\multirow{3}[2]{*}{GAN-based}} & pix2pix \cite{isola2017image} & $63.06$ & $63.07$ & $0.5392$ & $52.61$ & $0.9372$ \\
              & CycleGAN \cite{zhu2017unpaired} & $66.57$ & $66.86$ & $0.4582$ & $53.52$ & $0.9507$ \\
              & HiFaceGAN \cite{yang2020hifacegan} \dag & $66.94$ & $67.00$ & $0.4937$ & $55.29$ & $0.9922$ \\
        \midrule
        \multicolumn{1}{c}{\multirow{6}[2]{*}{Diffusion-based}} & ILVR \cite{choi2021ilvr} & $56.94$ & $57.19$ & $0.7163$ & $48.51$ & $0.9072$ \\
              & SR3 \cite{saharia2022image} & $61.20$ & $60.41$ & $0.7958$ & $51.84$ & $0.9189$ \\
              & SDEdit \cite{meng2021sdedit} & $64.91$ & $64.72$ & $0.6462$ & $49.16$ & $0.9117$ \\
              & EGSDE \cite{zhao2022egsde} & $64.26$ & $64.26$ & $0.6008$ & $48.75$ & $0.9081$ \\
              & T2V \cite{nair2023t2v} \dag & $66.48$ & $66.46$ & $0.2713$ & $59.70$ & $0.9920$ \\
              & Ours (proposed) & $\bm{70.19}$ & $\bm{70.28}$ & $\bm{0.1827}$ & $\bm{65.75}$ & $\bm{0.9994}$ \\
        \bottomrule
        \end{tabular}%
    \label{tab:main-oulu}%
\end{table*}%

\begin{table*}[t]
    \centering
    \vspace{-3mm}
    \caption{The performance of the compared methods on Large-HFE. \dag\  denotes methods specially designed for heterogeneous face translation. The best results are highlighted in \textbf{bold}.}
    \vspace{-3mm}
    \begin{tabular}{cl|cc|ccc}
        \toprule
        \multicolumn{2}{c|}{\multirow{2}[2]{*}{Method}} & \multicolumn{2}{c|}{Performance} & \multicolumn{3}{c}{Image Quality} \\
        \multicolumn{2}{c|}{} & Acc ($\%$) & F1 ($\%$) & LPIPS (↓) & PSNR (↑) & SSIM(↑) \\
        \midrule
        \multicolumn{2}{c|}{Without translation} & $67.07$ & $67.16$ & $0.5364$ & $50.88$ & $0.9417$ \\
        \midrule
        \multicolumn{1}{c}{\multirow{3}[2]{*}{GAN-based}} & pix2pix \cite{isola2017image} & $67.17$ & $67.02$ & $0.5657$ & $52.69$ & $0.9589$ \\
              & CycleGAN \cite{zhu2017unpaired} & $72.49$ & $72.52$ & $0.4801$ & $54.03$ & $0.9677$ \\
              & HiFaceGAN \cite{yang2020hifacegan} \dag & $69.68$ & $69.75$ & $0.4142$ & $55.26$ & $0.9962$ \\
        \midrule
        \multicolumn{1}{c}{\multirow{6}[2]{*}{Diffusion-based}} & ILVR \cite{choi2021ilvr} & $68.67$ & $69.05$ & $0.6053$ & $50.88$ & $0.9420$ \\
              & SR3 \cite{saharia2022image} & $65.86$ & $65.54$ & $0.5687$ & $54.59$ & $0.9691$ \\
              & SDEdit \cite{meng2021sdedit} & $69.88$ & $69.76$ & $0.5534$ & $51.11$ & $0.9439$ \\
              & EGSDE \cite{zhao2022egsde} & $70.78$ & $70.91$ & $0.5704$ & $50.94$ & $0.9423$ \\
              & T2V \cite{nair2023t2v} \dag & $72.59$ & $72.87$ & $0.3965$ & $56.12$ & $0.9830$ \\
              & Ours (proposed) & $\bm{74.90}$ & $\bm{74.91}$ & $\bm{0.3878}$ & $\bm{58.98}$ & $\bm{0.9963}$ \\
        \bottomrule
        \end{tabular}%
        \vspace{-0.2cm}
    \label{tab:main-HFE}%
\end{table*}%

\vspace{-0.2cm}
\subsection{Implementation Details}
The proposed method is implemented by Pytorch 1.8, and the model training and image sampling procedure is processed on NVIDIA GeForce 4090 card. All images are first cropped to $256\times256$ using  the face detector \cite{bulat2017far} and then both NIR\&VIS facial expression datasets are divided into training set and testing set. The training set is used to train the backbone model. The backbone conditional score matching network is implemented by U-Net. During training, the batch size is set to $16$ and the learning rate is set to $1\times10^{-5}$. After the model is ready, we can sample images using our method from CFEE and KDEF for NIR FER performance experiment and from testing set for VIS-NIR translation quality experiment. The hyperparameter in sampling process $\lambda_h$ is set to 100 and $\lambda_f$ is set to 0.5. For each experiment setting, the datasets are randomly divided into five parts for five independent experiments and averaged results are reported in the following part. 
\vspace{-0.2cm}
\section{Results and Analysis}
In this section, we first compare the proposed method with the baselines in terms of NIR FER performance and VIS-NIR translation quality. Then we conduct ablation study to analyze the effectiveness of each component in the proposed method. Finally, we visualize the generated NIR images to have qualitative evaluation of the compared methods.
\vspace{-0.2cm}
\subsection{Comparison with State-of-the-art Methods}

\begin{figure*}[htbp]
    \centering
     \vspace{-5mm}
    \includegraphics[width=0.99\textwidth]{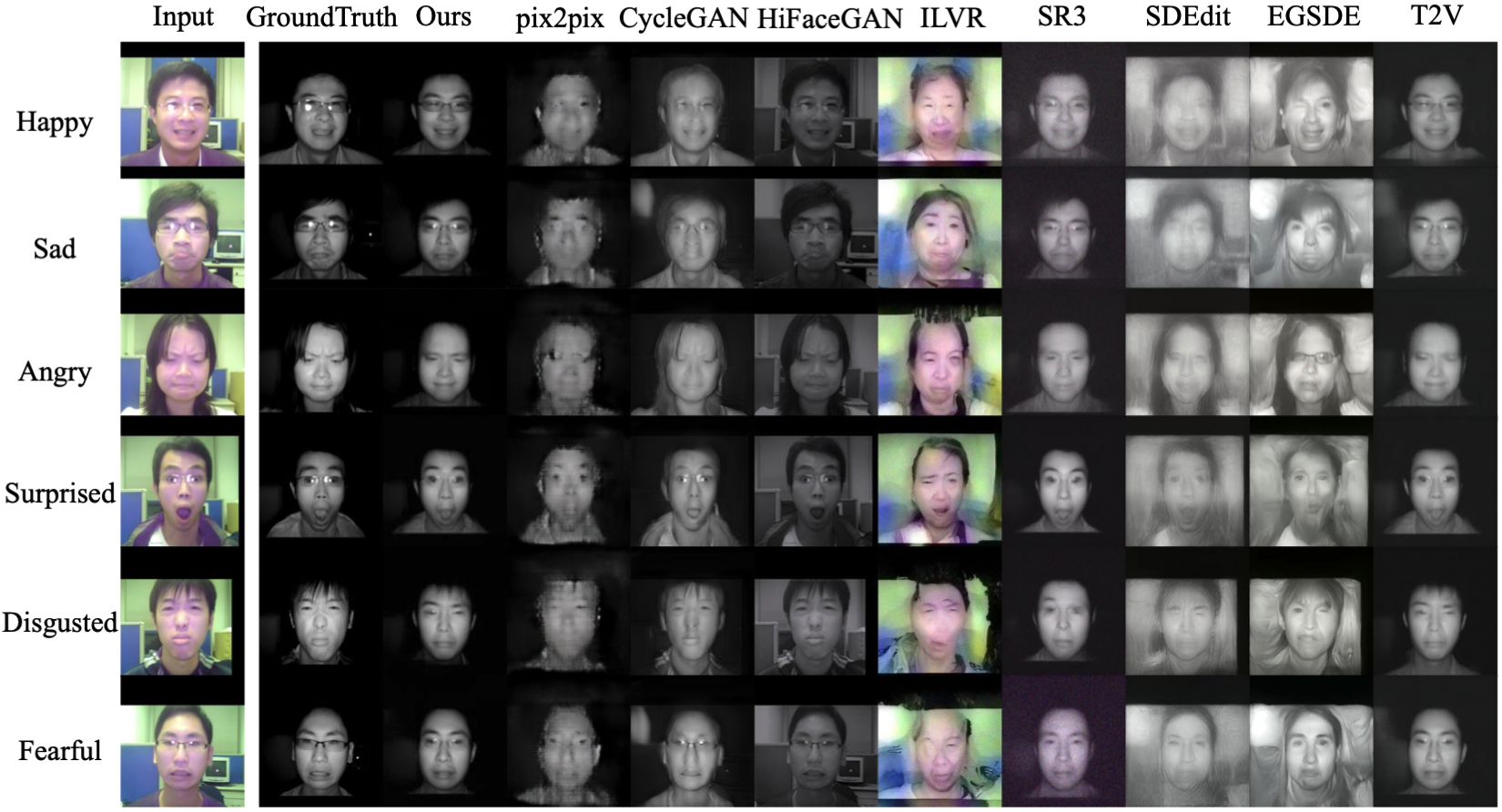} 
    \vspace{-3mm}
    \caption{Visualization of generated samples from the compared methods on Oulu-CASIA.}
    \label{fig:generated-samples-oulu}
    \vspace{-4mm}
\end{figure*}

To evaluate the proposed method, we compare it with the state-of-the-art methods in the settings of improved NIR FER performance and VIS-NIR translation quality on Oulu-CASIA and Large-HFE. The quantitative results are shown in Table \ref{tab:main-oulu} and Table \ref{tab:main-HFE} respectively. From the results, we have the observations as the followings.

(1) The proposed method shows consistent advantages over the state-of-the-art baselines in terms of improved NIR FER performance and VIS-NIR translation quality on both datasets. Compared to the best baseline, the proposed method achieves the improvement of accuracy by $3.25\%$ and $2.31\%$, and F1 score by $3.28\%$ and $2.04\%$ on Oulu-CASIA and Large-HFE. This indicates that the proposed method can translate VIS facial expression images to NIR modality with satisfying image quality and joint training with the translated images can effectively enhance the performance of NIR FER benchmark. This is mainly due to introduction of task-specific energy functions and conditional score matching, which guide the model to focus on more subtle yet key facial representations that are related to facial expression.

(2) Among the baselines, T2V and HiFaceGAN tend to show consistent effectiveness in image quality on both datasets. This is because T2V and HiFaceGAN are specially designed for heterogeneous face translation, and thus can generate more realistic NIR images of the facial details than the other baselines. CycleGAN also produces relatively competitive results, since the detailed original VIS image is considered by the GAN-based network. The other diffusion-based models, including ILVR, SDEdit, and EGSDE, perform not as expected, which is mainly due to the ignorance of details in the original image information in the translation process.

\vspace{-0.2cm}
\subsection{Ablation Study}
\begin{table}[htbp]
    \centering
    \vspace{-0.2cm}
    \caption{The ablation study results on Oulu-CASIA and Large-HFE. The best results are highlighted in \textbf{bold}.}
    \vspace{-0.3cm}
    \begin{tabular}{l|cc|cc}
        \toprule
        \multicolumn{1}{c|}{\multirow{2}[1]{*}{Method}} & \multicolumn{2}{c|}{Oulu-CASIA} & \multicolumn{2}{c}{Large-HFE} \\
            & Acc ($\%$) & F1 ($\%$) & Acc ($\%$) & F1 ($\%$) \\
        \midrule
        Baseline & $62.55$ & $62.75$ & $69.12$ & $69.43$ \\
        Cond   & $65.76$ & $66.06$ & $71.99$ & $72.17$ \\
        Cond+$\gE_h$ & $67.78$ & $68.98$ & $72.39$ & $72.49$ \\
        Cond+$\gE_f$ & $68.24$ & $68.57$ & $73.59$ & $73.53$ \\
        Cond+$\gE_h$+$\gE_f$ & $\bm{70.19}$ & $\bm{70.28}$ & $\bm{74.90}$ & $\bm{74.91}$ \\
        \bottomrule
    \end{tabular}%
    \vspace{-0.2cm}
    \label{tab:ablation}%
\end{table}%

The proposed method mainly consists of two parts, i.e., the conditional score-matching network and the task-specific energy guidance. In order to analyze the effectiveness of each component, we conduct ablation study on Oulu-CASIA and Large-HFE under the following settings:
\begin{itemize}
    \item \textbf{Baseline}: Basic SBDM (unconditional score-matching network and no task-specific energy function).
    \item \textbf{Cond}: Conditional score-matching network.
    \item \textbf{Cond+$\gE_h$}: Conditional score-matching network with energy function $\gE_h$ in Eq. (\ref{eq:landmark-energy}).
    \item \textbf{Cond+$\gE_f$}: Conditional score-matching network with energy function $\gE_f$ in Eq. (\ref{eq:high-frequency-energy}).
    \item \textbf{Cond+$\gE_h$+$\gE_f$}: The proposed method that combines all the components.
\end{itemize}

The ablation study results are shown in Table \ref{tab:ablation}. Observations can be drawn from the results as follows.

(1) The baseline of unconditional score-matching network and no task-specific energy function performs the worst, which is due to a lack of detailed facial representations and lack of knowledge guidance in the reverse process.

(2) Adding conditional score-matching network significantly improves the performance. This is because the added condition of the source facial expression image provides the score-matching network with key facial details to generate more realistic NIR facial expressions. Adding two task-specific energy functions further improves the performance respectively, which can guide the generation process toward more realistic NIR facial expressions with satisfying quality.

(3) The proposed method that jointly combines the above components achieves the best performance. This indicates that both the task-specific energy guidance and the conditional score-matching network are effective to improve the performance of NIR FER benchmark.

\subsection{Visualization of Generated Samples}

For qualitative evaluation results, we present the translated samples of the proposed method and the baselines on Oulu-CASIA in Fig. \ref{fig:generated-samples-oulu}. As shown in the figure, the proposed method can generate NIR images with satisfying quality, while the baselines cannot. Methods including ILVR, pix2pix, SDEdit, and EGSDE  generate human facial images with varying degrees of distortion on both datasets. Among them, ILVR performs the worst, since even fails to translate VIS images to NIR-like images. All other methods can meet the basic requirements of VIS-NIR translation. However, they still have some disadvantages compared to our proposed method. Images generated from CycleGAN and HiFaceGAN still suffer from black square artifact on Large-HFE, and on Oulu-CASIA, CycleGAN misses expression-related facial details (as in \textit{Angry}), while HiFaceGAN remains the background in VIS images. SR3 and T2V images have a paler color tone compared to real NIR images, which can be seen more clearly on Oulu-CASIA. Furthermore, SR3 fails to generate images of the same person on Large-HFE in some cases (As in \textit{Angry}, \textit{Surprised}, \textit{Disgusted}). In conclusion, the qualitative analysis results are consistent with the quantitative results, which further verifies the effectiveness of the proposed method.

%% file: data/conclusion.tex
\vspace{-0.2cm}
\section{Conclusion}
In this paper, we propose a novel framework named NFER-SDE on top of the score-based diffusion models to translate VIS facial expression samples to the NIR domain, so that the performance of NIR FER task can be improved with the translated training images. The framework consists of the energy functions of task-specific guidance and the conditional score-matching network, which guide the score-based diffusion models to capture subtle yet key facial representations for FER. Extensive experiments on two existing NIR FER datasets demonstrate the effectiveness of our method.